\RequirePackage{fix-cm}
\documentclass[twocolumn]{svjour3}          
\smartqed  
\usepackage{graphicx}
\usepackage{ulem}
\usepackage[round]{natbib}
\usepackage{url}
\usepackage[ruled,vlined]{algorithm2e}
\usepackage{bbm}

\usepackage{xcolor}

\SetKwInput{KwInput}{Input}
\SetKwInput{KwOutput}{Output}

\begin{document}

\title{A deep learning approach to clustering visual arts
}


\author{Giovanna Castellano         \and
        Gennaro Vessio 
}


\institute{G. Castellano \and G. Vessio \at
              Department of Computer Science \\
              University of Bari ``Aldo Moro'', Italy \\
              \email{gennaro.vessio@uniba.it}           
}

\date{Received: date / Accepted: date}

\maketitle

\begin{abstract}
Clustering artworks is difficult for several reasons. On the one hand, recognizing meaningful patterns based on domain knowledge and visual perception is extremely hard. On the other hand, applying traditional clustering and feature reduction techniques to the highly dimensional pixel space can be ineffective. To address these issues, in this paper we propose {\sf DELIUS}: a DEep learning approach to cLustering vIsUal artS. The method uses a pre-trained convolutional network to extract features and then feeds these features into a deep embedded clustering model, where the task of mapping the input data to a latent space is jointly optimized with the task of finding a set of cluster centroids in this latent space. Quantitative and qualitative experimental results show the effectiveness of the proposed method. {\sf DELIUS} can be useful for several tasks related to art analysis, in particular visual link retrieval and historical knowledge discovery in painting datasets.
\end{abstract}

\section{Introduction}
Cultural heritage, especially visual art, is of inestimable importance for the cultural, historical, and economic growth of our societies. In recent years, due to technological improvements and the drastic drop in costs, a large-scale digitization effort has been made which has led to a growing availability of large digitized artwork collections. Notable examples include WikiArt\footnote{\url{https://www.wikiart.org}} and the MET collection.\footnote{\url{https://www.metmuseum.org/art/collection}} This availability, coupled with recent advances in pattern recognition and computer vision, has opened up new opportunities for computer science researchers to assist the art community with tools to analyze and support further understanding of visual arts. Among other things, a deeper understanding of visual arts has the potential to make them accessible to a wider population, both in terms of fruition and creation, ultimately supporting the spread of culture \citep{NCAA}.

The ability to recognize meaningful patterns in visual artworks is intrinsically related to the domain of human perception \citep{leder2004model}. The recognition of the stylistic and semantic attributes of an artwork, in fact, originates from the composition of the color, texture, and shape features visually perceived by the human eye, and can be influenced by previous historical knowledge (or lack thereof). Furthermore, human aesthetic perception depends on subjective experience and on the emotions the artwork evokes in the observer. This is why this perception can be extremely difficult to conceptualize. Nevertheless, representation learning approaches, especially those upon which deep learning models are based \citep{bengio2013representation,lecun2015deep}, may be the key to success in extracting useful features from low-level color and texture features. These approaches are already helpful for various tasks related to art analysis, from period estimation, e.g.~\citep{strezoski2017omniart}, to style classification, e.g.~\citep{cetinic2018fine}. 

Although there is a growing body of knowledge on applying pattern recognition and computer vision algorithms to this domain in a supervised fashion, see for example \citep{karayev2013recognizing,crowley2014search,garcia2020contextnet}, very little work has been done in the unsupervised setting. The supervised approach is really useful for solving different classification and retrieval tasks. However, it requires considerable effort to accurately annotate artworks, which, as mentioned above, can be influenced by some subjectivity. Furthermore, it is less suited to support more in-depth analysis, as the machine is typically constrained to solve only a specific task. 

Conversely, having a model that can cluster artworks without depending on hard-to-collect labels or subjective knowledge can be useful for many other domain applications. It can be used to support art experts in discovering trends and influences from one art movement to another, i.e.~in performing ``historical'' knowledge discovery. Similarly, the model can be used to discover different periods in the production of the same artist. It can support interactive browsing of online collections by finding visually linked artworks, thus performing a ``visual'' link retrieval. It can help curators organize permanent or temporary exhibitions based on visual similarities rather than just historical motivations. Finally, the model can help experts classify contemporary art, which cannot be provided with rich annotations, and, similarly, it can help find out which artworks have most influenced the work of current artists.

To this end, in this paper, we propose {\sf DELIUS}: a DEep learning approach to cLustering vIsUal artS. The method is based on using a pre-trained deep convolutional neural network (CNN), that is DenseNet121 \citep{huang2017densely}, as an unsupervised feature extractor, and then using a deep embedded clustering (DEC) model \citep{xie2016unsupervised} to perform the final clustering. The choice of this fully deep pipeline was motivated by the difficulty of applying traditional clustering algorithms and feature reduction techniques to the highly dimensional input pixel space, especially when dealing with very complex artistic images. The effectiveness of the method was evaluated on a subset of the previously mentioned widely used WikiArt collection. It is worth noting that this paper extends our previous work in this direction \citep{ICPR}, by introducing a more refined model, and more experiments and ablation studies on a much larger dataset.

The rest of the paper is organized as follows. Section~2 deals with related work. Section~3 describes the proposed method. Sections~4 and~5 present and discuss the experimental setup and the results obtained. Section~6 concludes the paper and outlines directions for further research on the topic.

\section{Related work}
\label{related}
Traditionally, automated art analysis has been done using hand-crafted features fed into traditional machine learning algorithms, e.g.~\citep{shamir2010impressionism,6460929,carneiro2012artistic,khan2014painting}. Unfortunately, despite the encouraging results of feature engineering techniques, early attempts soon stalled due to the difficulty of gaining explicit knowledge about the attributes to associate with a particular artist or artwork. This difficulty arises because this knowledge typically depends on an implicit and subjective experience that a human expert might find difficult to verbalize. In fact, experts draw their judgments based on various factors, notably the historical context of the work, as well as the understanding of the metaphor behind what is immediately perceived \citep{saleh2016toward}. Furthermore, art experts, as well as inexperienced enthusiasts, can have subjective reactions to the stylistic properties of an artwork, as emotions can influence their aesthetic perception \citep{cetinic2019deep}.

In contrast, several successful applications in many computer vision tasks have demonstrated the effectiveness of representation learning versus feature engineering techniques in extracting meaningful patterns from complex raw data; seminal papers such as \citep{lecun1989backpropagation} and \citep{krizhevsky2012imagenet} are well known to the community. One of the main reasons for the success of deep neural network models in solving tasks too difficult for classical pipelines is the availability of large datasets with human annotations, such as ImageNet \citep{deng2009imagenet}. A model built on these data, in fact, often provides a sufficiently general knowledge of the ``visual world'' that can be profitably transferred to specific visual domains, in particular the artistic one. This provided an opportunity to tackle historically difficult tasks in the art domain, mitigating the need for manually extracting features, and exploiting higher-level, easier-to-collect labels, such as the time of production or the school of painting the artworks belong to, to apply data-driven learning strategies.

One of the first successful attempts to apply deep neural networks in this context was the research presented by \cite{karayev2013recognizing}, which shows how a CNN pre-trained on PASCAL VOC can be quite effective in attributing the correct school of painting to an artwork. Since then, many articles have been devoted to the use of deep learning techniques based on single-input, e.g.~\citep{van2015toward,chen2019recognizing}, or multi-input models, e.g.~\citep{strezoski2017omniart,garcia2020contextnet}, to solve artwork attribute prediction tasks based on visual features.

Another task often faced by the research community working in this field is finding objects in artworks. Indeed, art historians are often interested in finding out when a specific object first appeared in a painting or how the representation of an object evolved over time. A pioneering work in this context has been the research of \cite{crowley2014search}. They proposed a system that, given an input query, retrieves positive training samples by crawling Google Images on the fly. These are then processed by a pre-trained CNN and used together with a pre-computed pool of negative features to learn a real-time classifier. Since then, many other works have explored this direction further, e.g.~\citep{cai2015cross,westlake2016detecting}, also focusing on weakly supervised approaches \citep{gonthier2018weakly} or near duplicate detection tasks \citep{shen2019discovering}. The main issue encountered in this research is the so-called \textit{cross-depiction} problem, which is the problem of recognizing visual objects regardless of whether they are photographed, painted, drawn, etc. The variance between photos and artworks is greater than both domains when considered alone, so classifiers usually trained on traditional photographic images typically encounter difficulties when used on painting images, due to the domain shift. This is still an open issue in the research community, and an approach based on artistic to photo-realistic translation has recently been proposed to try to fill this gap \citep{tomei2019art2real}.

Another task that has attracted attention in the art domain is the development of a machine capable of imitating human creativity to some extent. Traditional literature on computational creativity has developed systems for the generation of art based on the involvement of human artists in the generation process. More recently, the advent of the Generative Adversarial Network paradigm \citep{goodfellow2014generative} has allowed researchers to develop systems that do not put humans in the loop but make use of previous human artworks in the learning process. This is consistent with the assumption that even human experts use prior experience and their knowledge of past art to develop their own creativity. Some popular models, in particular \citep{elgammal2017can} and \citep{tan2018improved}, have been developed that generate images rated as highly plausible by human experts.

In recent times, a research direction that has sparked increasing interest is the one that combines computer vision with natural language processing techniques to provide a unified framework for solving multi-modal retrieval tasks. In this view, the system is asked to find an artwork based on textual comments describing it and vice versa. Notable works in this direction are the research of \cite{garcia2018read} and that of \cite{cornia2020explaining}.

Most of the existing literature reports the use of deep learning-based solutions that require some form of supervision. Conversely, very little work has been done from an unsupervised perspective. In \citep{barnard2001clustering}, the authors proposed a clustering approach to artistic images by exploiting textual descriptions with natural language processing. \cite{spehr2009image}, on the other hand, applied a computer vision approach to the problem of clustering paintings using traditional hand-crafted features. \cite{gultepe2018predicting} applied an unsupervised feature learning method based on $k$-means to extract features which were then fed into a spectral clustering algorithm for the purpose of grouping paintings. In \citep{MTAP}, we have recently proposed a method for finding visual links among paintings in a completely unsupervised way. The method relies solely on visual attributes learned automatically by a deep pre-trained model and finds similarities between paintings in a nearest neighbor fashion. Unfortunately, while it is effective in addressing the link retrieval problem, this approach is not suitable for finding clusters in the feature space.

\section{Proposed method}
\label{method}
Clustering is one of the fundamental tasks in Machine Learning. It is notoriously difficult, mainly due to the lack of supervision on how to guide the search for patterns in the data and in evaluating what the algorithm finds. In particular, since its appearance, $k$-means has been widely used due to its ease of implementation and effectiveness. However, especially in a complex image domain, applying $k$-means may not be feasible. On the one hand, clustering with traditional distance measures in the highly dimensional raw pixel space is well known to be ineffective. Furthermore, as noted earlier, extracting meaningful feature vectors based on domain-specific knowledge can be extremely difficult when dealing with artistic data. On the other hand, applying well-known dimensionality reduction techniques, such as PCA, to the original space or a manually engineered feature space, can ignore possible nonlinear relationships between the original input and the latent space, thus decreasing clustering performance. To get around these difficulties, we propose a fully deep learning pipeline. 

\begin{figure}
    \centering
    \includegraphics[width=.8\columnwidth]{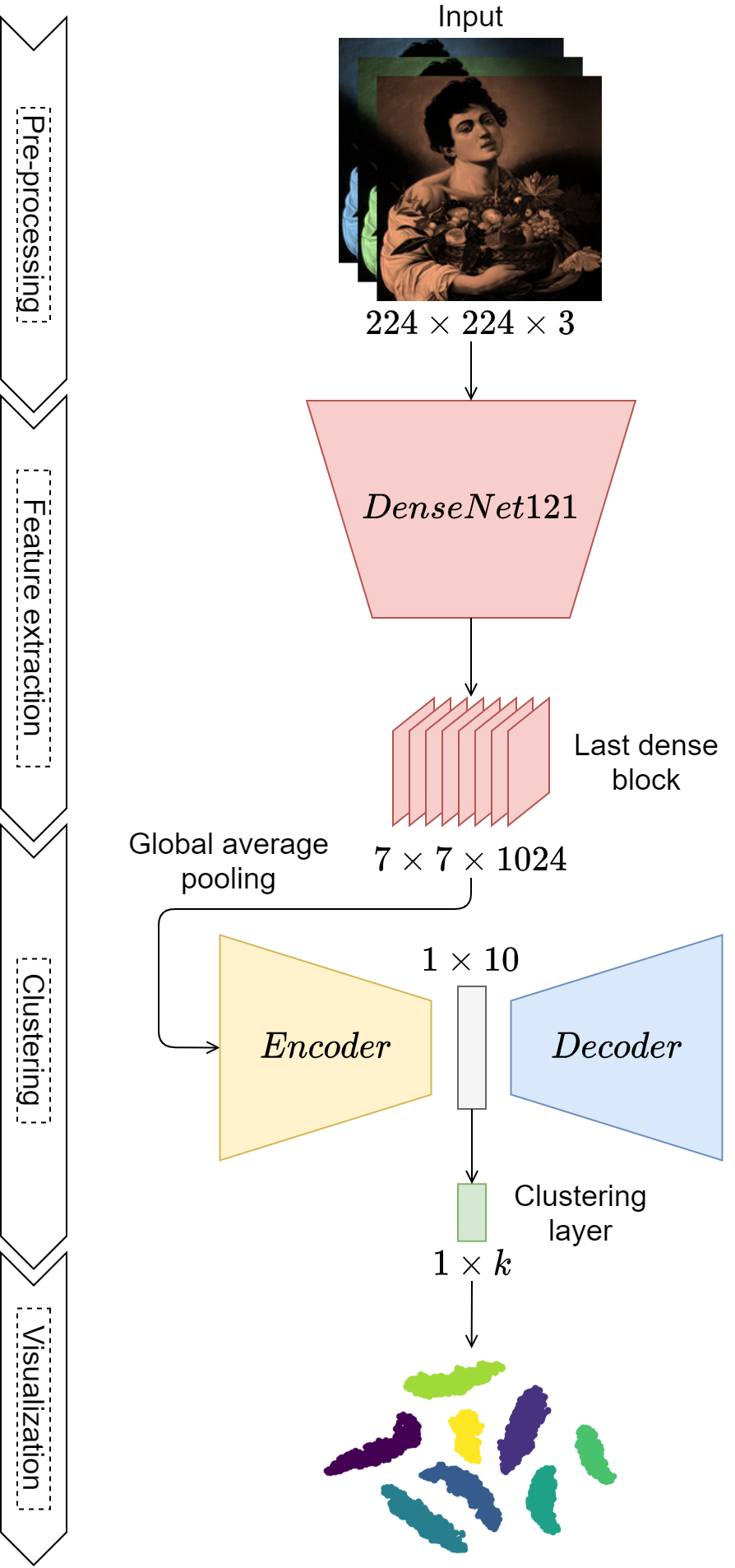}
    \caption{Schema of the proposed method. Three-channel images follow an information processing flow, undergoing some transformations, namely pre-processing, feature extraction, feature reduction through nonlinear embedding and cluster assignment, and a final reduction in two dimensions.}
    \label{fig:schema}
\end{figure}

{\sf DELIUS}, which is schematized in Fig.~\ref{fig:schema}, consists of the following steps:
\begin{enumerate}
    \item Artwork images are first pre-processed; 
    \item Pre-processed images are fed into DenseNet121 to extract visual features; 
    \item The global average pooled features are provided as input to a deep embedded clustering model to perform  clustering;
    \item The embedded features are finally projected in two dimensions with t-SNE for visualization purposes.
\end{enumerate}
Algorithm \ref{algo} summarizes the main steps. Details of each step are provided in the following subsections. 

\subsection{Pre-processing}
Each RGB image is first resized to $224 \times $224, which is the input size normally accepted by DenseNet121, and its pixel values are scaled between 0 and 1, as is usually done when using convolutional neural networks.

\subsection{Feature extraction}
The Dense Convolutional Network (DenseNet) family of models connects each layer to each other layer in a feed-forward fashion \citep{huang2017densely}. While traditional CNNs with $L$ layers have $L$ direct connections, one between each layer and the next one, DenseNets have $L(L + 1)/2$ connections. For each layer, the feature maps of all previous layers are used as input, and the own feature maps are used as input to all subsequent layers. DenseNets are often considered the logical extension of the previous well-known ResNet \citep{he2016deep}: the key difference is the use of concatenation instead of addition in cross-layer connections. The name \textit{DenseNet} derives from the fact that the dependency graph between the variables becomes rather dense: the last layer of the chain is densely connected to all the previous layers. The metaphor is that each layer receives the ``knowledge gathered'' from all the previous layers and together they collectively contribute to the final output. The main components that form a DenseNet are \textit{dense blocks} and \textit{transition layers}. The former define how the inputs and outputs are chained; the latter control the number of channels so that it is kept small.

DenseNets have several advantages over other state-of-the-art architectures: they alleviate the vanishing gradient problem, strengthen feature propagation, encourage feature reuse, and substantially reduce the number of parameters. Like any CNN, the network is able to build a hierarchy of visual features, starting with simple edges and shapes in earlier layers to higher-level concepts like complex objects and shapes as the layers go deeper. This approach is suitable for obtaining high-level semantic representations from the initial artwork images without the need for any supervision. To obtain these features, we use the common practice of extracting features from the last dense block, which returns 1024 $7 \times 7$ feature maps.

The feature maps are finally global average pooled to get a more compact one-dimensional vector of size 1024. As usual, this fairly simple operation is performed to significantly reduce the feature dimensionality.

\subsection{Clustering}
In recent years, a deep clustering paradigm has emerged that exploits the ability of deep neural networks to find complex nonlinear relationships among data for clustering purposes \citep{xie2016unsupervised,guo2017deep,yang2017towards}. The idea is to jointly optimize the task of mapping the input data to a lower dimensional space and the task of finding a set of cluster centroids in this latent space. Deep clustering is being used with very promising results in several real domains, e.g.~\citep{bhowmik2018deep,lu2019audio}; however, it is usually applied directly to the original input. In this paper, we 
leverage the ability of a deep network like DenseNet121 to extract meaningful and more compact features from very complex artistic images before clustering them.

The global average pooled features from the previous step are provided as input to a deep embedded clustering model, such as the DEC model proposed by \cite{xie2016unsupervised}. Basically, DEC is based on an autoencoder and a so-called clustering layer connected to the embedded layer of the autoencoder. Autoencoders are neural networks that learn to reconstruct their input using an encoder and a decoder in combination.
The encoder transforms the data with a nonlinear mapping $\phi:X\rightarrow Z$, where $X$ is the input space and $Z$ is a smaller hidden latent space. The decoder learns to reconstruct the input using this latent representation, $\psi:Z\rightarrow X$. This is done in a completely unsupervised way, with no knowledge of a target variable.
In addition to the input layer, which depends on the specific input shape, and which in our case is equal to 1024, the encoder size has been set as in \citep{xie2016unsupervised} to 500-500-2000-10 hidden units, where 10 is the size of the latent embedded space. The embedded features are then reshaped and propagated through the decoder part of the network, which mirrors the encoder hyperparameters in reverse order and restores the embedded features back to the original input space.
To learn the nonlinear mappings $\phi$ and $\psi$, the autoencoder parameters are updated by minimizing a classic mean squared reconstruction loss:
\begin{equation}\label{eq:r_loss}
    \mathcal{L}_r = \frac{1}{n}\sum_{i=1}^{n}\left\|\mathbf{x}'_i - \mathbf{x}_i \right\|^2 = \frac{1}{n}\sum_{i=1}^{n}\left\|\psi\left( \phi \left( \mathbf{x}_i \right) \right) - \mathbf{x}_i \right\|^2
\end{equation}
where $n$ is the cardinality of the set of input data, $\mathbf{x}_i$ is the $i$-th input sample, $\mathbf{x}'_i$ its reconstruction, and $\|\cdot \|$ is the Euclidean distance.

The clustering layer attached to the embedded layer of the autoencoder assigns the embedded features of each sample to a cluster. Given an initial estimate of the nonlinear mapping $\phi : X \rightarrow Z$, and the $k$ centroids of an initial clustering $\{\mathbf{c}_j \in Z\}_{j=1}^k$, the clustering layer maps each embedded point, $\mathbf{z}_i=\phi(\mathbf{x}_i)$, to a cluster centroid, $\mathbf{c}_j$, using a cluster assignment distribution $Q$ based on Student's $t$-distribution:
\begin{equation}
\label{q_dist}
q_{ij} = \frac{\left( 1 + \left \| \mathbf{z}_i - \mathbf{c}_j \right \|^2 \right)^{-1}}{\sum_{j'} \left( 1 + \left \| \mathbf{z}_i - \mathbf{c}_{j'} \right \|^2 \right)^{-1}}
\end{equation}
where $q_{ij}$ represents the membership probability of $\mathbf{z}_i$ of belonging to cluster $j$; in other words, it can be seen as a soft assignment. Membership probabilities are used to calculate an auxiliary target distribution $P$:
\begin{equation}
\label{p_dist}
p_{ij} = \frac{q_{ij}^2/f_{i}}{\sum_{j'} q_{ij'}^2 / f_{j'}}
\end{equation}
where $f_{j}=\sum_i q_{ij}$ are soft cluster frequencies. Clustering is done by minimizing the Kullback-Leibler (KL) divergence between $P$ and $Q$:
\begin{equation}
\label{eq:c_loss}
    \mathcal{L}_c = KL(P \parallel Q) = \sum_i \sum_j p_{ij} \log \left( \frac{p_{ij}}{q_{ij}} \right)
\end{equation}
In practice, the $q_{ij}$'s provide a measure of the similarity between each data point and the different $k$ centroids. Higher values for $q_{ij}$ indicate more confidence in assigning a data point to a particular cluster. The auxiliary target distribution is designed to place greater emphasis on the data points assigned with greater confidence while normalizing the loss contribution of each centroid. Then, by minimizing the divergence between the membership probabilities and the target distribution, the network improves the initial estimate by learning from previous high-confidence predictions.

The overall training works in two steps. In the first step, the autoencoder is trained to learn an initial set of embedded features, minimizing the reconstruction loss defined in Eq.~\ref{eq:r_loss}. After this pre-training phase, the learned features are used to initialize the cluster centroids $\mathbf{c}_j$ using traditional $k$-means. Finally, the decoder part of the model is abandoned and embedded feature learning and clustering are jointly optimized by minimizing only the cluster assignment loss defined in Eq.~\ref{eq:c_loss}. The overall weights are updated using backpropagation. It is worth noting that, to avoid instability, $P$ is not updated on every iteration using only a batch of data, but using all embedded points every $t$ iterations. The training procedure stops when the change in cluster assignments between two consecutive updates is below a given threshold $\delta$. 

It is worth underlining that ``training'' here means the process of optimizing the reconstruction of the original input, in the case of the autencoder, and the search for cluster centroids, in the case of clustering. Both tasks do not require any form of supervision.

\begin{algorithm}[t]
\SetAlgoLined
\KwInput{A given dataset 
$\mathcal{D}$ of $n$ images; 
number of clusters $k$}
Preprocess $\mathcal{D}$;\\
For each 
image $I \in \mathcal{D}$, extract a feature vector $\mathbf{x}_i$ using DenseNet121 plus global average pooling;\\
Initialize the autoencoder by minimizing Eq.~\ref{eq:r_loss} to get initial embeddings $\{ \mathbf{z}_i \in Z \}_{i=1}^n$;\\
Initialize cluster assignment with $k$-means to get initial cluster centroids $\{ \mathbf{c}_j \}_{j=1}^k$;\\
\While{not converged}{
Compute $q_{ij}$ using Eq.~\ref{q_dist}\;
Compute $p_{ij}$ using Eq.~\ref{p_dist}\;
Update the encoder and cluster assignment by minimizing Eq.~\ref{eq:c_loss}\; 
}
Further reduce the dimensionality of $Z$ with t-SNE;
\\
\KwResult{Each image $I$ is assigned to a cluster centroid $\mathbf{c}_j$}
\caption{{\sf DELIUS}}
\label{algo}
\end{algorithm}

\subsection{Visualization}
After training, the initial artwork images lie in the 10-dimensional embedded feature space. These features can then be used to fit a t-SNE representation to display data in a reduced space, for qualitative evaluation purposes. t-distributed Stochastic Neighbor Embedding is a well-known technique for nonlinear dimensionality reduction, which converts the similarities between the data points into joint probabilities and tries to minimize the KL divergence between the joint probabilities of the lower dimensional embeddings and the higher dimensional data \citep{van2008visualizing}.

It is worth noting that, contrary to other clustering approaches, the reduced feature space will differ depending on the number of clusters, as the proposed method simultaneously minimizes image reconstruction and cluster assignment. Therefore, changing the number of clusters will lead to different arrangements of the data in the space induced by the embedding.

\section{Experimental setting}
This Section describes the dataset we used, implementation details, other clustering approaches with which we compared our method, and evaluation metrics.

\subsection{Dataset}
Several datasets have been used and proposed by the research community working on digitized art \citep{NCAA}. Some of them, such as People-Art \citep{westlake2016detecting}, are actually intended for object detection tasks; others, such as SemArt \citep{garcia2018read}, were designed for multi-modal retrieval tasks. WikiArt (formerly known as WikiPaintings) is currently one of the largest online collections of digitized paintings available, and has been a frequent choice for dataset creation in many recent studies and has contributed to several art-related research projects. The artworks collected cover a wide range of periods, with a particular focus on the art of the last century. The dataset is constantly growing and includes not only paintings but also sculptures, sketches, posters, and so on. As of this writing, WikiArt includes nearly $170,000$ artworks.

We used the data already scraped and downloaded from WikiArt by \cite{tan2016ceci},\footnote{\url{https://github.com/cs-chan/ArtGAN}} for a total of $78,978$ digitized artworks, including not only paintings but also drawings and illustrations. The artworks span a wide range of historical periods (from Early Renaissance to Pop Art) and genres (from portraits to landscapes). To obtain a fairly balanced dataset between the classes, for later evaluation, and to avoid excessive fragmentation, we grouped the artworks into 8 well-known macro-periods. These are shown in Table \ref{tab:data}. Furthermore, to evaluate the clustering performance by genre and not just by style, we also distinguished the artworks by genre (see Table \ref{tab:data-genre}). Note that since the genre is not provided for all the artworks in the dataset, they add up to $64,524$.

\begin{figure*}
    \centering
    \includegraphics[width=.8\textwidth]{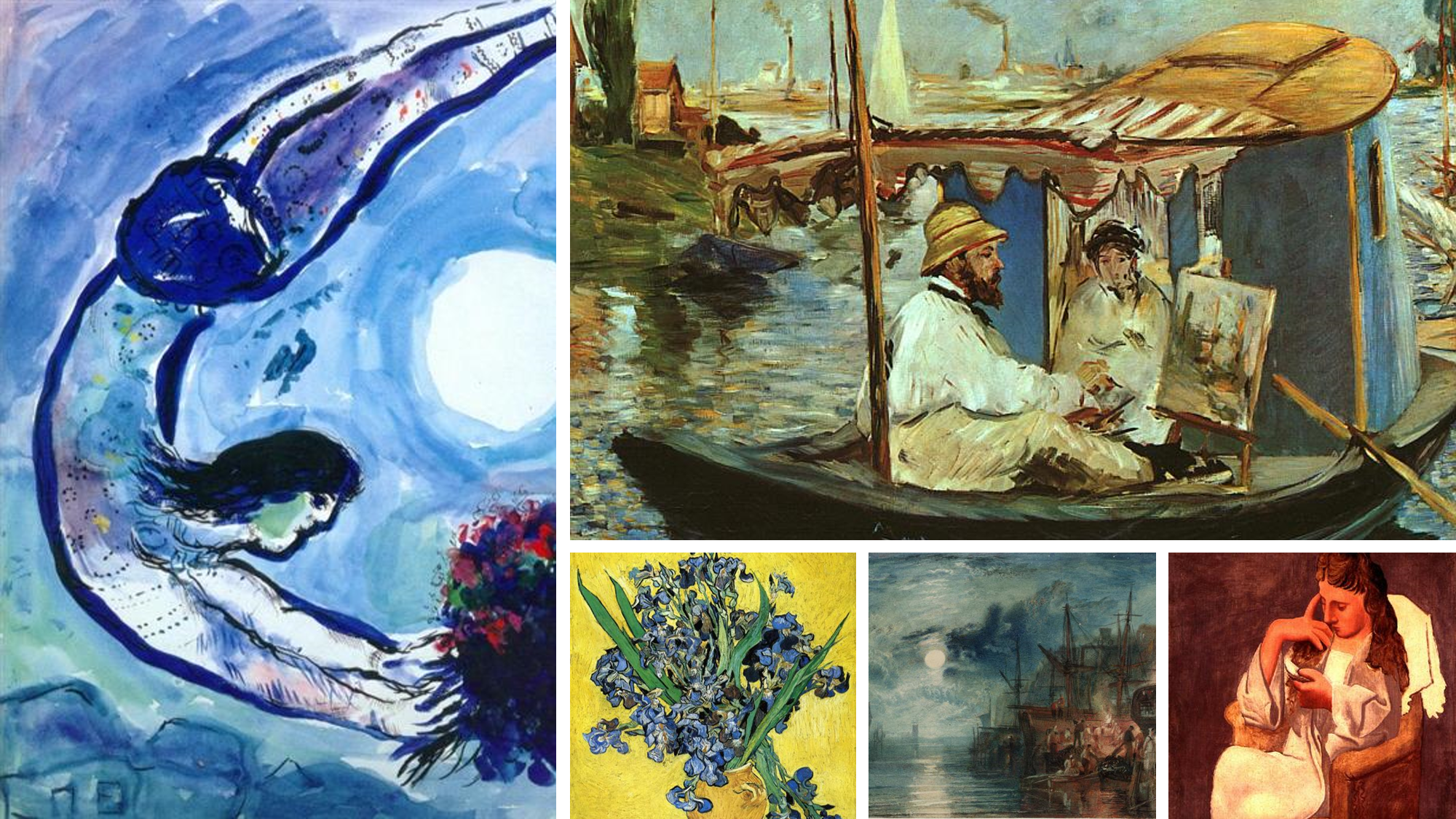}
    \caption{Examples of artworks from WikiArt. From top to bottom, from left to right: ``Acrobat with Bouquet'' by Marc Chagall; ``Monet in his Studio Boat'' by Edouard Manet; ``Still Life with Irises'' by Vincent van Gogh; ``Shields, on the River Tyne'' by William Turner; ``Woman Reading'' by Pablo Picasso.}
    \label{fig:samples}
\end{figure*}

Artwork images vary in size but, as noted earlier, they were resized and normalized in the pre-processing stage. Sample images are shown in Fig.~\ref{fig:samples}.

\begin{table}
\centering
\caption{Art movements of our dataset, along with the artwork count. The coverage period of these movements is from the 15th century to the present day and, as it is notoriously not easy to define them rigorously, they sometimes overlap.}
\label{tab:data}
\begin{tabular}{lr}
\hline\noalign{\smallskip}
Movement & \# artworks\\
\noalign{\smallskip}\hline\noalign{\smallskip}
Renaissance & $6,575$\\
Baroque and Rococo & $6,329$\\
Romanticism & $6,945$\\
Realism & $10,609$\\
Impressionism & $13,016$\\
Post-Impressionism & $11,892$\\
Expressionism and Cubism & $13,371$\\
Modern and Contemporary Art & $10,241$\\
\noalign{\smallskip}\hline
\end{tabular}
\end{table}

\begin{table}
\centering
\caption{Art genres of our dataset, along with the artwork count.}
\label{tab:data-genre}
\begin{tabular}{lr}
\hline\noalign{\smallskip}
Genre & \# artworks\\
\noalign{\smallskip}\hline\noalign{\smallskip}
Abstract paintings & $5,446$\\
Cityscapes & $4,535$\\
Genre paintings & $10,342$\\
Illustrations & $1,883$\\
Landscapes & $13,102$\\
Nude paintings & $1,853$\\
Portraits & $14,055$\\
Religious paintings & $6,482$\\
Sketches and studies & $4,012$\\
Still lifes & $2,814$\\
\noalign{\smallskip}\hline
\end{tabular}
\end{table}

\subsection{Implementation details}
The experiments were performed on an Intel Core i5 equipped with the NVIDIA GeForce MX110, with dedicated memory of 2 GB. As a deep learning framework, we used TensorFlow 2.0 and the Keras API.

Regarding the hyper-parameter setting, the autoencoder was initially pre-trained with mini-batch size 256, and using Adam for 200 epochs with a recommended learning rate of 0.001, $\beta_1$ of 0.9, $\beta_2$ of 0.999 and $\epsilon$ of $1 \times 10^{-8}$. The weights were randomly initialized from a normal distribution with a mean of zero and a standard deviation of 0.01, as suggested in \citep{xie2016unsupervised}. For initialization of the cluster centroids, we used traditional $k$-means with 20 restarts, choosing the best solution. Finally, the deep embedded clustering model was trained using Adam with the same set of parameters as before, but setting the convergence threshold $\delta$ to 0.001 and the update interval $t$ to 140.

\subsection{Other clustering approaches}
Since no work in the literature uses deep learning for clustering paintings, we compared our proposed method with other clustering approaches to the same problem to assess whether it provides a better solution. In particular, we considered three alternative strategies:
\begin{itemize}
    \item $k$-means on the original pixel space previously reduced by a PCA to decrease the extremely high dimensionality (from now on-wards {\sf PCA+$k$-means});
    \item $k$-means on the embedded feature space resulting from the pre-training phase of the autoencoder, before training it together with the clustering layer (from now on-wards {\sf AE+$k$-means});
    \item The deep convolutional embedding clustering model proposed in our previous preliminary work \citep{ICPR}, which introduced some slight changes to the one proposed by \cite{guo2017deep} (from now on-wards {\sf DCEC}). In short, the encoder and decoder are made up of convolutional layers and the model is fed directly with the three-channel images.
\end{itemize}

It is worth noting that, since our goal was not to propose a new generic deep clustering model but a specific methodology that would work effectively in the domain of computational analysis of art, we did not extend the comparison to other methods similar in philosophy to {\sf DCEC}.

\subsection{Evaluation metrics}
Since clustering is unsupervised, we do not know \textit{a priori} which is the best grouping of paintings. Furthermore, since even two artworks by the same artist could have been produced in different stylistic periods, it is very difficult to assign a precise label to a given painting, thus providing a form of accurate supervision over cluster assignment. For this reason, for clustering evaluation, we mainly used two standard internal metrics, namely the silhouette coefficient \citep{rousseeuw1987silhouettes} and the Calinski-Harabasz index \citep{calinski1974dendrite}, which are based on the model itself. The silhouette coefficient is defined for each sample and is calculated as follows:
\begin{equation}
    SC = \frac{b-a}{\max(a,b)}
\end{equation}
where $a$ is the average distance between a data point and all other points in the same cluster, and $b$ is the average distance between a data point and all other points in the nearest cluster. The final score is obtained by averaging over all data points. The silhouette coefficient is between $-1$ and $1$, representing the worst and best possible value respectively. Values close to $0$ indicate overlapping clusters. The Calinski-Harabasz index is the ratio of the sum of between-cluster dispersion and inter-cluster dispersion for all clusters. More precisely, for a dataset $\mathcal{D}$ of size $n$, which has been partitioned into $k$ clusters, the index is defined as:
\begin{equation}
    CHI = \frac{\mathrm{tr}(B_k)}{\mathrm{tr}(W_k)} \times \frac{n - k}{k - 1}
\end{equation}
where $\mathrm{tr}(B_k)$ is the trace of the between-cluster dispersion matrix, being $B_k = \sum_{j=1}^k n_j (\mu_j - \mu_\mathcal{D}) (\mu_j - \mu_\mathcal{D})^\top$, and $\mathrm{tr}(W_k)$ is the trace of the within-cluster dispersion matrix defined as $ W_k = \sum_{j=1}^k \sum_{x \in C_j} (x - \mu_j) (x - \mu_j)^\top$, where $C_j$ is the set of points in cluster $j$, $\mu_j$ the center of cluster $j$, $\mu_\mathcal{D}$ the center of $\mathcal{D}$, and $n_j$ the cardinality of cluster $j$. The Calinski-Harabasz index is not bounded within a given interval. It is worth remarking that, except for {\sf PCA+$k$-means}, both $SC$ and $CHI$ were computed in the space induced by the embedding. 
Indeed, these metrics help assess whether the deep model produced adequate representations from the original high-dimensional data when provided with an appropriate objective function.

The above metrics are based on internal criteria. However, note that the eight periods into which the dataset we used can be broken down provide a form of ground truth. Furthermore, artworks can be further divided into 10 classes if we divide them by genre. In this way, we can also use external criteria such as the unsupervised clustering accuracy \citep{cai2010locally}, which is widely used in the unsupervised setting:
\begin{equation}
    ACC = \max_m \frac{\sum_{i=1}^{n} \mathbbm{1} \{y_i = m(c_i)\}}{n}
\end{equation}
where $y_i$ is the ground-truth label, $c_i$ is the cluster assignment, and $m$ varies over all possible one-to-one mappings between clusters and labels. $ACC$ differs from the usual classification accuracy in that it uses the mapping function $m$ to find the best mapping between the cluster assignment $c$ and the ground truth $y$. This mapping is necessary because an unsupervised algorithm can use a different label than the actual ground truth label to represent the same cluster. Another popular external metric, useful when knowledge of the ground truth is available, is the normalized mutual information \citep{vinh2010information}:
\begin{equation}
NMI = \frac{MI(U, V)}{\mathrm{mean}(H(U), H(V))}
\end{equation}
where $MI(U, V) = \sum_{i=1}^{|U|} \sum_{j=1}^{|V|} \frac{|U_i \cap V_j|}{N}\log\left(\frac{N|U_i \cap V_j|}{|U_i||V_j|}\right)$ is the classic mutual information calculated between the cluster assignments $U$ and the ground truth labels $V$, normalized by the average entropy $H$ of both. Values close to $0$ indicate two largely independent assignments, while values close to $1$ indicate significant agreement.

Finally, we also drew qualitative observations on the cluster assignments provided by the method based on the t-SNE visualization. 
Note that other reduction strategies similar to t-SNE can be used for the same purpose, such as Uniform Manifold Approximation and Projection (UMAP) \citep{mcinnes2018umap}.

\begin{figure*}
    \includegraphics[width=\textwidth]{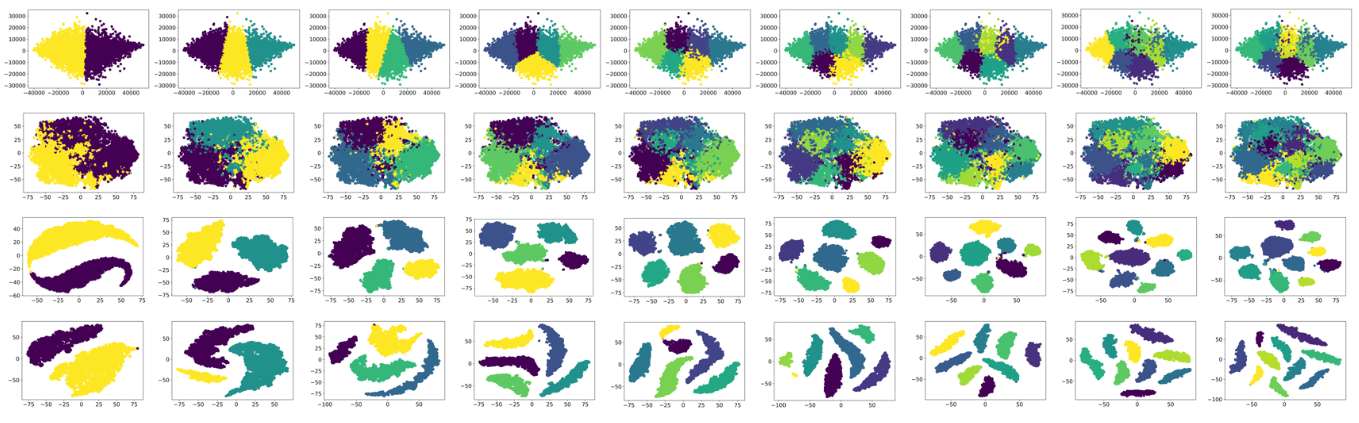}
    \caption{From top to bottom: bi-dimensional visualizations of the comparison between {\sf DELIUS} (fourth row) and {\sf PCA+$k$-means} (first row), {\sf AE+$k$-means} (second row) and {\sf DCEC} (third row), varying the number of clusters $k$ from 2 to 10. Note that, for better visualization, these views were obtained by plotting the same stratified random sample of 10\% of the overall data.}
    \label{fig:t-sne}
\end{figure*}

\begin{figure*}
    \centering
    \includegraphics[width=.99\textwidth]{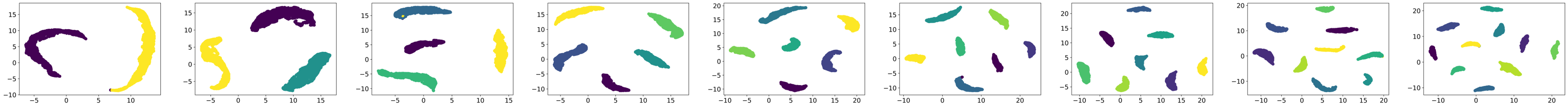}
    \caption{Bi-dimensional visualization of the clusters found by {\sf DELIUS} with UMAP. For the sake of comparison with t-SNE, the plots were obtained on the same fraction of data as in Fig.~\ref{fig:t-sne}.}
    \label{fig:umap}
\end{figure*}

\section{Results}
We carried out several experiments to evaluate the effectiveness of the proposed method. The first experiment was devoted to evaluating the effectiveness of the method in clustering the overall data. A second experiment was dedicated to showing the effectiveness of the method in grouping artworks created by the same artist. In a third experiment, we compared the proposed solution with the other clustering approaches to the same problem. Finally, we show the results of some ablation studies aimed at adjusting the final system.

\subsection{Clustering the overall data}
\begin{figure}
    \centering
    \includegraphics[width=\columnwidth]{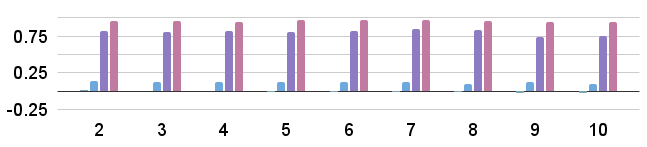}
    \includegraphics[width=\columnwidth]{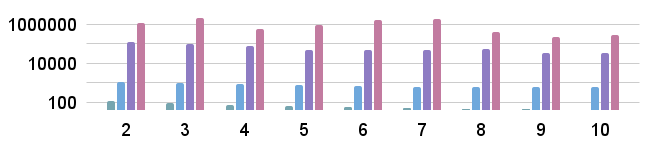}
    \caption{Comparison between {\sf DELIUS} (magenta) and {\sf PCA+$k$-means} (cyan), {\sf AE+$k$-means} (blue) and {\sf DCEC} (purple) in terms of silhouette coefficient (top) and Calinski-Harabasz index (bottom).}
    \label{fig:soa_comparison_uns}
\end{figure}

\begin{figure}
    \centering
    \includegraphics[width=\columnwidth]{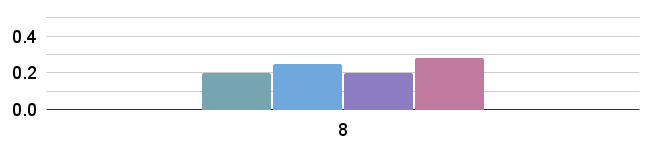}
    \includegraphics[width=\columnwidth]{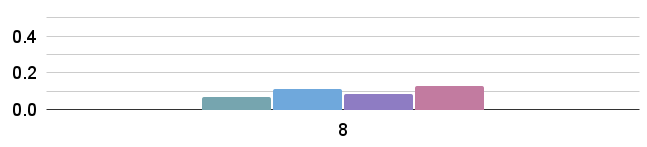}
    \caption{Comparison between {\sf DELIUS} (magenta) and {\sf PCA+$k$-means} (cyan), {\sf AE+$k$-means} (blue) and {\sf DCEC} (purple) in terms of clustering accuracy (top) and normalized mutual information (bottom) by style.}
    \label{fig:soa_comparison_style}
\end{figure}

\begin{figure}
    \centering
    \includegraphics[width=\columnwidth]{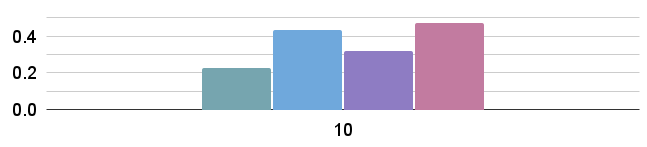}
    \includegraphics[width=\columnwidth]{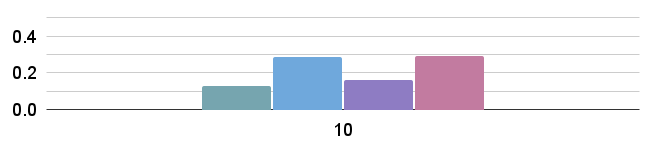}
    \caption{Comparison between {\sf DELIUS} (magenta) and {\sf PCA+$k$-means} (cyan), {\sf AE+$k$-means} (blue) and {\sf DCEC} (purple) in terms of clustering accuracy (top) and normalized mutual information (bottom) by genre.}
    \label{fig:soa_comparison_genre}
\end{figure}

\begin{figure}
    \centering
    \includegraphics[width=\columnwidth]{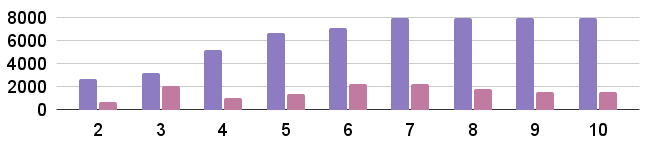}
    \caption{Comparison between {\sf DELIUS} (magenta) and {\sf DCEC} (purple) in terms of number of iterations needed to converge.}
    \label{fig:soa_comparison_it}
\end{figure}

Figures \ref{fig:t-sne}--\ref{fig:soa_comparison_it} show the clustering performance of the proposed method over the entire dataset by varying the number of clusters $k$. Note that, to avoid redundancies, the figures also show the results obtained with other methods, which will be discussed in Sect.~5.3. We varied $k$ between 2 and 10, which is the grouping suggested by the ten different genres to which the artworks in the dataset belong. Observing the two-dimensional visualization given by t-SNE, it can be seen that in all cases {\sf DELIUS} is able to create well-separated clusters (Fig.~\ref{fig:t-sne}).
To assess whether the method finds well-separated clusters even when a different reduction technique is used, we also experimented with UMAP instead of t-SNE, which provided even better concentrated clusters (see Fig.~\ref{fig:umap}).
These results are quantitatively confirmed by the silhouette coefficient, whose value is always close to 1, and by the Calinski-Harabasz index, which has very high values for all partitions (Fig.~\ref{fig:soa_comparison_uns}).

As regards clustering accuracy and normalized mutual information, {\sf DELIUS} obtained unsatisfactory results if we consider the eight historical classes into which the dataset was divided (Fig.~\ref{fig:soa_comparison_style}).
In particular, the accuracy of the model was $0.28 \pm 0.01$ at the $95\%$ confidence level, slightly higher than the top second accuracy of $0.25 \pm 0.01$ obtained by {\sf AE+$k$-means}. In all cases, $NMI$ shows lower results than $ACC$.
However, significantly higher results were obtained when considering the genre ground truth (Fig.~\ref{fig:soa_comparison_genre}). In fact, although the clustering accuracy by genre does not exceed 50\% ($0.47 \pm 0.01$ at the $95\%$ confidence level, higher than the top second accuracy of $0.43 \pm 0.01$ obtained by {\sf AE+$k$-means}), it should be noted that in a classic supervised setting this would be a 10-class classification problem. Again, $NMI$ behaves worse than $ACC$, although better than the previous evaluation based on style. These results suggest that the model tends to look at content rather than stylistic features to group paintings.

\begin{figure*}
    \centering
    \includegraphics[width=.65\textwidth]{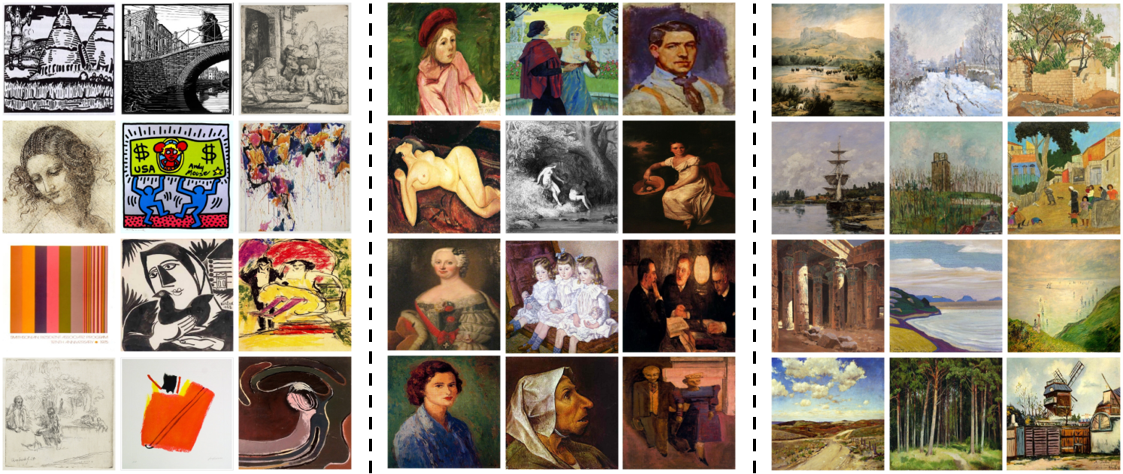}
    \caption{Sample images from the clusters found by {\sf DELIUS} when applied to the overall dataset with $k=3$.}
    \label{fig:3-clusters}
\end{figure*}

\begin{figure*}
    \centering
    \includegraphics[width=.9\textwidth]{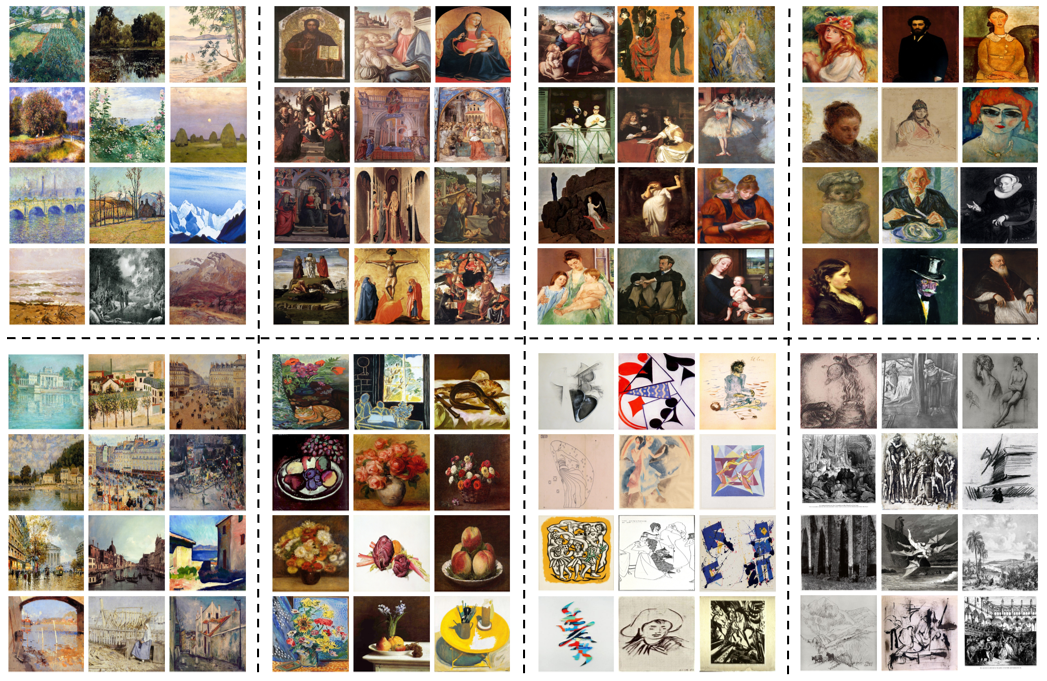}
    \caption{Sample images from the clusters found by {\sf DELIUS} when applied to the overall dataset with $k=8$.}
    \label{fig:8-clusters}
\end{figure*}

A more complete assessment was made by analyzing the composition of the clusters and visualizing artworks belonging to each cluster. To this end, in Fig.~\ref{fig:3-clusters} and Fig.~\ref{fig:8-clusters} we show some exemplary images from the clusters found by {\sf DELIUS} when $k=3$ and $k=8$. It can be seen that in the case of three clusters (Fig.~\ref{fig:3-clusters}), {\sf DELIUS} separated the dataset into three macro-categories: drawings and illustrations in a cluster; portraits and, more generally, paintings depicting people in a second cluster; finally, landscapes and cityscapes in the third cluster. When eight clusters are considered (Fig.~\ref{fig:8-clusters}), the model further fragments the data into groups showing homogeneous characteristics: the drawings and illustrations are separated into two distinct clusters; the works depicting people are divided into portraits only, genre paintings, and religious paintings; finally, a clearer separation was made between landscapes, cityscapes and still lifes. As the number of clusters grows, {\sf DELIUS} separates artworks into smaller and smaller groups that share visual similarities.
Interestingly, the clusters found tend to reflect some known artistic influences and connections between artists. For example, sketches and studies by Duerer and Da Vinci, known to have contributed significantly to Renaissance,\footnote{\url{https://en.wikipedia.org/wiki/Renaissance}} appear in the same cluster. Likewise, the model tends to group religious paintings\footnote{\url{https://en.wikipedia.org/wiki/Religious_art}} by artists such as Pontormo and van Eyck, as well as cityscapes and landscapes by Monet and Manet.\footnote{\url{https://en.wikipedia.org/wiki/Impressionism}}

This qualitative assessment confirms that the model created by {\sf DELIUS} takes 
into account semantic features, that is, those relating to the subject and genre of the artwork, rather than just stylistic properties. This can be explained considering that the deep clustering model uses high-level of abstraction features extracted from a deep pre-trained CNN as an input, which typically involve more complex shapes and objects rather than low-level features. Furthermore, unlike supervised approaches, where the same deep CNN can be used (see Sect.~5.4), after feature extraction the deep clustering model is not forced to learn useful features for style classification tasks, but independently explores high-level visual similarities between artworks to group them. In other words, when a clear loss function based on a style ground truth is adopted, the model is forced to learn useful features to minimize that loss. Instead, in an unsupervised setting, the model is free to group artworks independently of such loss and tends to consider content-based features. These findings are consistent with previous work showing how visual features can be used to find similarities between artists \citep{saleh2016toward,MTAP} and how visual features from the same semantic domain overlap while appearing separate from other domains \citep{garcia2018read,cornia2020explaining}.
Furthermore, as the results we have obtained with the supervised approach also demonstrate (Sect.~5.4), even when the system has a clear loss function to minimize, the period prediction is difficult, as more classical works can exhibit futuristic and pioneering features, while modern works can draw inspiration and revive the classic style. In addition, there is usually confusion between similar styles \citep{strezoski2017omniart}. Indeed, 
precisely separating artistic styles is still challenging.

\begin{figure*}
    \centering
    \includegraphics[width=.7\textwidth]{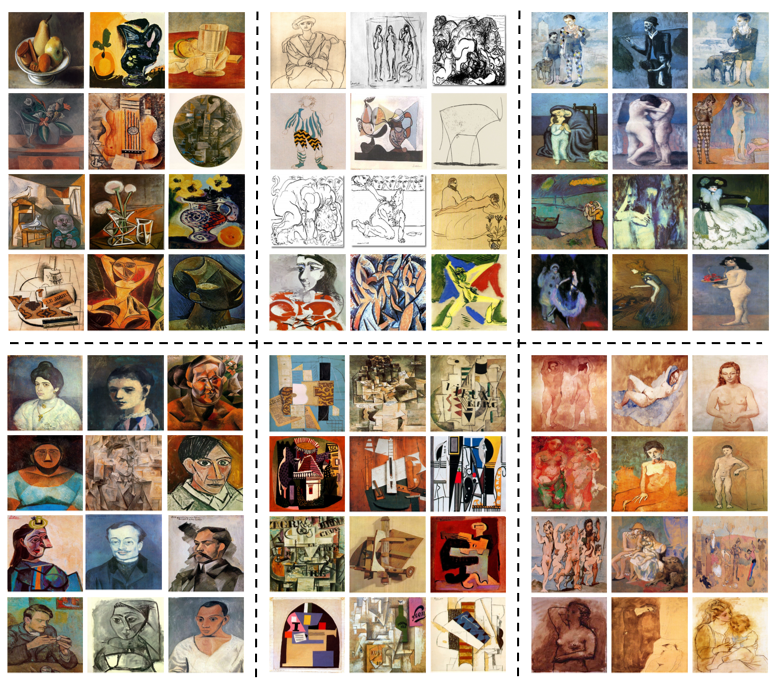}
    \caption{Sample images from the clusters found by {\sf DELIUS} among Picasso's artworks when $k=6$.}
    \label{fig:picasso}
\end{figure*}

Clearly, the proposed method is far from being perfect, especially considering that, like any clustering algorithm, the method is forced to group data, so a data point can appear in one cluster because it does not have enough similarities to any other cluster. However, the ability of {\sf DELIUS} to group objects regardless of how they are depicted (in a realistic or abstract way, and so on) makes it capable of mimicking, to some extent, the human semantic understanding of art. This may be a step towards solving the well-known cross-depiction problem, since, as suggested by \citep{cai2015cross,cai2015beyond}, a candidate solution is not to learn the specificity of each representation, but to learn the abstraction that the different representations share so that they can be recognized regardless of their depiction. Abstract styles in particular, such as Expressionism and Cubism, have always posed serious challenges, as representations of objects and subjects can show strong individuality and therefore less generalizable patterns.

\subsection{Clustering a single artist}
We also experimented with a sub-sample of the dataset comprising the works of a single artist. This was done to evaluate the effectiveness of the proposed method in the search for meaningful clusters within the production of the same artist.

In particular, we run the proposed method on the 762 artworks made by Pablo Picasso provided by the dataset we used, as Picasso is known for having a prolific production with different series, genres, and periods of works. We tested different values of $k$, from 2 to 8. In all cases, well-separated clusters were found, with the best value for the silhouette coefficient and the Calinski-Harabasz index when $k = 6$, where $SC \approx 0.97$ and $CHI \approx 6.5 \times 10^4$. Figure \ref{fig:picasso} shows sample images from these 6 clusters. Again, the model created by {\sf DELIUS} has shown a tendency to group together works with notable visual similarities, related to the subject matter and the content of the work: one cluster is clearly related to still lifes; another with sketches and studies; another cluster with portraits (including self-portraits); two clusters appear to be related to the so-called ``blue'' and ``rose'' periods of the author; finally, one cluster is related to the late neoclassicist and surrealist works. 

When only a small dataset of a single artist is considered, {\sf DELIUS} finds it easier to distinguish paintings by style as well. 
Picasso's major stylistic phases have been described by a variety of scholars, writers, and critics; and {\sf DELIUS} seems to have found some well-known groups such as the aforementioned ``blue'' and ``rose'' period, as well as the later works.\footnote{\url{https://en.wikipedia.org/wiki/Pablo_Picasso}}
The model also manages to mitigate the cross-depiction problem, as it places semantically related works, such as portraits and still lifes, in the same groups, despite their stylistic depiction. The Cubist still lifes, for example, are not separate from the more realistic ones.

\subsection{Comparison with other clustering approaches}
Figures \ref{fig:t-sne}--\ref{fig:soa_comparison_it} also show the results of the comparison between the four methods we have tested on the overall data from a qualitative and quantitative point of view. Figure \ref{fig:t-sne} shows the bi-dimensional visualizations of the clusters obtained with the methods. These clearly outline how {\sf PCA+$k$-means} and {\sf AE+$k$-means} cannot find well-defined and separated clusters in the feature space. The data points appear randomly distributed and overlapping in the bi-dimensional space without any structure emerging. The behavior shown by {\sf PCA+$k$-means} was expected since applying traditional techniques to the original pixel space in such a complex domain is well known to be ineffective. The behavior of {\sf AE+$k$-means} was also expected. Although reduced by CNN processing, the resulting feature space still has a high dimensionality and $k$-means is unable to find well-formed clusters in this space. On the contrary, as noted earlier, {\sf DELIUS} is capable of separating data in well-formed clusters in all cases. The strategy of further reducing the feature space through the autoencoder and optimizing embedding learning with clustering appears to be effective for grouping artworks. Likewise, {\sf DCEC}, which is based on a similar deep clustering principle, is able to find well-formed clusters.

These qualitative results are reflected by the quantitative evaluation shown in Fig.~\ref{fig:soa_comparison_uns}. The values for the silhouette coefficient and the Calinski-Harabasz index remain quite stable and high regardless of the number of clusters $k$ for {\sf DELIUS}. The values for these metrics shown by {\sf DCEC} are also good although lower than the proposed method. This confirms the effectiveness of using a deep pre-trained model like DenseNet to extract features before learning the embedding. Furthermore, it should be noted that being performed directly on the initial images, the {\sf DCEC} training is much more computationally demanding, as it requires many more iterations to converge (Fig.~\ref{fig:soa_comparison_it}). In contrast, {\sf PCA+$k$-means} and {\sf AE+$k$-means} show very poor performance, with the metric values progressively decreasing as $k$ increases, confirming their ineffectiveness to solve the clustering task. 

Finally, with regard to the external metrics (Figs.~\ref{fig:soa_comparison_style}-\ref{fig:soa_comparison_genre}), no model is able to obtain satisfactory results in relation to the time period ground truth. However, when considering the genre ground truth, both {\sf AE+$k$-means} and {\sf DELIUS} exceed 40\%. The better 
results achieved by {\sf AE+$k$-means} compared to {\sf DCEC}, although well-defined clusters are not obtained with the method, confirms that the features extracted by DenseNet may be more informative than those learned from scratch by the convolutional autoencoder.

\begin{figure*}
    \centering
    \includegraphics[width=.75\textwidth]{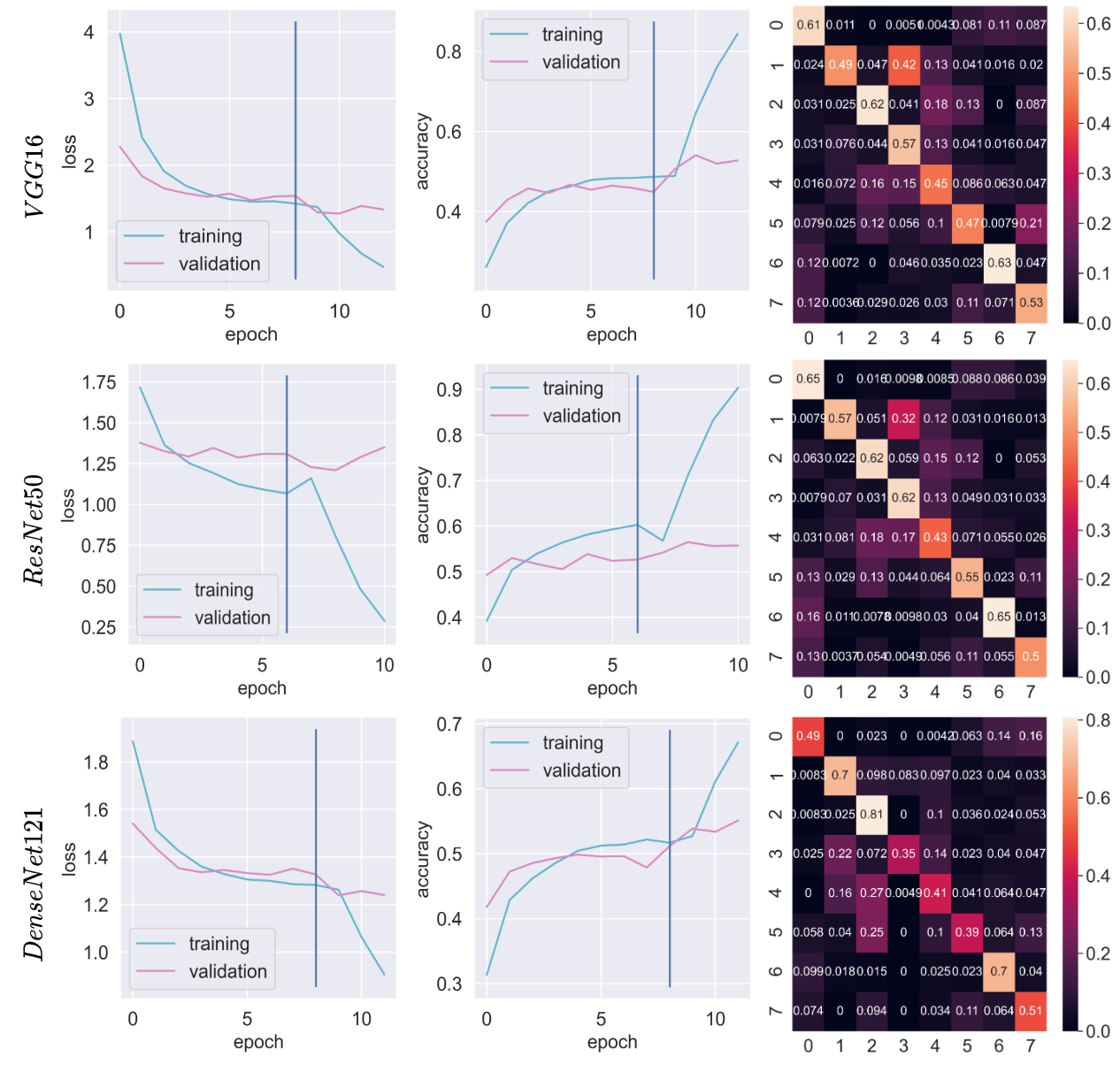}
    \caption{From top to bottom: the learning curves of loss and accuracy on the training and validation set and the normalized confusion matrix, for each network compared. The vertical lines indicate the start of fine-tuning.}
    \label{fig:ablation}
\end{figure*}

\subsection{Ablation study}
In this section, we report the results of some ablation studies we have done to choose the best neural network architecture to extract meaningful features from digitized paintings and drawings without supervision. To do this, we compared DenseNet121 with the popular VGG16 and ResNet50. VGG16 is a classic CNN architecture that adopts $3 \times 3$ convolution and $2 \times 2$ max pooling throughout the network and follows a standard scheme in which convolutional layers are interleaved with max pooling layers for a total of 16 weight layers \citep{simonyan2014very}. ResNet, on the other hand, uses a \textit{skip connection} strategy so that inputs can propagate faster between layers \citep{he2016deep}. As noted earlier, ResNet uses addition rather than concatenation in cross-layer connections.

To compare the three networks fairly, we trained them to solve the same classification task, namely the separation of the WikiArt dataset into the 8 stylistic classes of Table \ref{tab:data}. This was done to find the network that is best able to recognize some stylistic properties of the artworks. The evaluation was performed on a validation set obtained as a 10\% stratified fraction of the overall dataset. The three networks were trained using the same training strategy: each of them was equipped with a global average pooling layer on top of the convolutional base; a dropout layer with a dropout rate of 0.2; and a final output layer with softmax activation attached. Additionally, each network was trained to minimize a cross-entropy loss function using Adam with a learning rate of 0.001. We adopted the common practice of using the networks as feature extractors first, monitoring the validation loss and continuing propagation until early stopping with a patience of 2. Then, starting from weights pre-trained on ImageNet, we fine-tuned each network by unfreezing the last convolutional block while continuing propagation with a learning rate of $10^{-4}$ to prevent the previously learned weights from being destroyed. Again, we independently stopped each network training by monitoring the validation loss with a patience of 2.

The results obtained are shown in Fig.~\ref{fig:ablation}. They point out that none of the networks is able to achieve excellent results on WikiArt. However, it should be noted that this is an 8-class classification task, the random baseline of which is much lower. Furthermore, we have not implemented any other strategies to push these findings further, as this was not the main focus of our study. The learning curves show that ResNet50 tends to severely overfit the dataset very soon; while a slightly better behavior is shown by VGG16 and DenseNet121, with the latter performing better than the former. The lower number of parameters of DenseNet121 compared to the other two models makes it less prone to overfitting when fine-tuned on a dataset, WikiArt, which is significantly smaller than ImageNet. All in all, also considering that DenseNet121 is smaller in size and faster than the other two models, we preferred to use this network.

Finally, it is worth noting that fine-tuning was only used to help better assess network behavior on these data, but in our final model we only used DenseNet121 in an unsupervised way.

\section{Conclusion}

The contribution of this study was the achievement of new results in the automatic analysis of art, which is a very difficult task. Indeed, recognizing meaningful patterns in artworks based on domain knowledge and human visual perception is extremely difficult for machines. For this reason, the application of traditional clustering and feature reduction techniques to the highly dimensional pixel space has been largely ineffective. Automatic discovery of patterns in painting is desirable to relax the need for prior knowledge and labels, which are very difficult to collect in this field, even for an expert. 

To address these issues, in this paper we have proposed {\sf DELIUS}, a new fully unsupervised deep learning approach to clustering visual arts. The quantitative and qualitative experimental results show that {\sf DELIUS} was able to find well-separated clusters both when considering an overall dataset spanning different periods and when focusing on artworks produced by the same artist. 
In particular, from a qualitative point of view, it seems that {\sf DELIUS} is able to look not only at stylistic features to group artworks, but also especially at semantic attributes, relating to the content of the scene depicted. This abstraction capability appears to hold promise for addressing the well-known cross-depiction problem---which still poses a challenge to the research community---, pushing toward a better imitation of the human understanding of art.

As future work, 
we would like to study how the injection of some pieces of prior art knowledge, in a semi-supervised fashion, e.g.~\citep{ren2019semi}, can be advantageous for the proposed method and to better cluster the works of current artists. Furthermore, it is worth noting that, in an attempt to mimic human aesthetic perception, without using any prior knowledge, {\sf DELIUS} relies only on visual features to group artworks; however, artworks are characterized not only by their visual appearance but also by various other historical, social, and contextual factors that place them in a more complex scenario. A promising way to harness this knowledge is to encode (con)textual information of the artworks into a Knowledge Graph (KG) \citep{hogan2021knowledge} and use an appropriate representation of the nodes in the graph as an additional input to a deep learning model. Recent work has already moved in this direction \citep{garcia2020contextnet,vaigh2021gcnboost} and we are contributing by developing \textit{ArtGraph}, an artistic KG based on WikiArt and DBpedia \citep{castellano2022leveraging}. In the future, we would like to integrate \textit{ArtGraph} with {\sf DELIUS} to explore how the injection of (con)textual information of prior art is useful for generalizing beyond just the visual features, and thus for clustering novel art and to gain better semantic scene understanding and machine aesthetic perception.

\paragraph{Funding}
G.V.~acknowledges the financial support of the Italian Ministry of University and Research through the PON AIM 1852414 project.

\paragraph{Conflicts of interest}
The authors declare they have no conflict of interest.




\paragraph{Acknowledgement}
The authors would like to thank Vito Centoducati and Luigi Tedesco for their help with data curation and some 
comparative studies.

\bibliographystyle{spbasic}      
\bibliography{biblio}   

\begin{thebibliography}{55}
\providecommand{\natexlab}[1]{#1}
\providecommand{\url}[1]{{#1}}
\providecommand{\urlprefix}{URL }
\expandafter\ifx\csname urlstyle\endcsname\relax
  \providecommand{\doi}[1]{DOI~\discretionary{}{}{}#1}\else
  \providecommand{\doi}{DOI~\discretionary{}{}{}\begingroup
  \urlstyle{rm}\Url}\fi
\providecommand{\eprint}[2][]{\url{#2}}

\bibitem[{{Arora} and {Elgammal}(2012)}]{6460929}
{Arora} RS, {Elgammal} A (2012) Towards automated classification of fine-art
  painting style: A comparative study. In: Proceedings of the 21st
  International Conference on Pattern Recognition (ICPR 2012), pp 3541--3544

\bibitem[{Barnard et~al(2001)Barnard, Duygulu, and
  Forsyth}]{barnard2001clustering}
Barnard K, Duygulu P, Forsyth D (2001) Clustering art. In: Proceedings of the
  2001 IEEE Computer Society Conference on Computer Vision and Pattern
  Recognition. CVPR 2001, IEEE, vol~2

\bibitem[{Bengio et~al(2013)Bengio, Courville, and
  Vincent}]{bengio2013representation}
Bengio Y, Courville A, Vincent P (2013) Representation learning: A review and
  new perspectives. IEEE Transactions on Pattern Analysis and Machine
  Intelligence 35(8):1798--1828

\bibitem[{Bhowmik et~al(2018)Bhowmik, Gao, Young, and
  Ramanathan}]{bhowmik2018deep}
Bhowmik D, Gao S, Young MT, Ramanathan A (2018) Deep clustering of protein
  folding simulations. BMC Bioinformatics 19(18):47--58

\bibitem[{Cai et~al(2010)Cai, He, and Han}]{cai2010locally}
Cai D, He X, Han J (2010) Locally consistent concept factorization for document
  clustering. IEEE Transactions on Knowledge and Data Engineering
  23(6):902--913

\bibitem[{Cai et~al(2015{\natexlab{a}})Cai, Wu, Corradi, and
  Hall}]{cai2015cross}
Cai H, Wu Q, Corradi T, Hall P (2015{\natexlab{a}}) The cross-depiction
  problem: Computer vision algorithms for recognising objects in artwork and in
  photographs. arXiv preprint arXiv:150500110

\bibitem[{Cai et~al(2015{\natexlab{b}})Cai, Wu, and Hall}]{cai2015beyond}
Cai H, Wu Q, Hall P (2015{\natexlab{b}}) Beyond photo-domain object
  recognition: Benchmarks for the cross-depiction problem. In: Proceedings of
  the IEEE International Conference on Computer Vision Workshops, pp 1--6

\bibitem[{Cali{\'n}ski and Harabasz(1974)}]{calinski1974dendrite}
Cali{\'n}ski T, Harabasz J (1974) A dendrite method for cluster analysis.
  Communications in Statistics-theory and Methods 3(1):1--27

\bibitem[{Carneiro et~al(2012)Carneiro, da~Silva, Del~Bue, and
  Costeira}]{carneiro2012artistic}
Carneiro G, da~Silva NP, Del~Bue A, Costeira JP (2012) Artistic image
  classification: An analysis on the printart database. In: European Conference
  on Computer Vision, Springer, pp 143--157

\bibitem[{Castellano and Vessio(2021{\natexlab{a}})}]{ICPR}
Castellano G, Vessio G (2021{\natexlab{a}}) Deep convolutional embedding for
  digitized painting clustering. In: International Conference on Pattern
  Recognition (ICPR 2020), IEEE, to appear

\bibitem[{Castellano and Vessio(2021{\natexlab{b}})}]{NCAA}
Castellano G, Vessio G (2021{\natexlab{b}}) Deep learning approaches to pattern
  extraction and recognition in paintings and drawings: An overview. Neural
  Computing and Applications To appear

\bibitem[{Castellano et~al(2020)Castellano, Lella, and Vessio}]{MTAP}
Castellano G, Lella E, Vessio G (2020) Visual link retrieval and knowledge
  discovery in painting datasets. Multimedia Tools and Applications In press

\bibitem[{Castellano et~al(2022)Castellano, Digeno, Sansaro, and
  Vessio}]{castellano2022leveraging}
Castellano G, Digeno V, Sansaro G, Vessio G (2022) Leveraging knowledge graphs
  and deep learning for automatic art analysis. Knowledge-Based Systems
  248:108859

\bibitem[{Cetinic et~al(2018)Cetinic, Lipic, and Grgic}]{cetinic2018fine}
Cetinic E, Lipic T, Grgic S (2018) Fine-tuning convolutional neural networks
  for fine art classification. Expert Systems with Applications 114:107--118

\bibitem[{Cetinic et~al(2019)Cetinic, Lipic, and Grgic}]{cetinic2019deep}
Cetinic E, Lipic T, Grgic S (2019) A deep learning perspective on beauty,
  sentiment, and remembrance of art. IEEE Access 7:73694--73710

\bibitem[{Chen and Yang(2019)}]{chen2019recognizing}
Chen L, Yang J (2019) Recognizing the style of visual arts via adaptive
  cross-layer correlation. In: Proceedings of the 27th ACM International
  Conference on Multimedia, pp 2459--2467

\bibitem[{Cornia et~al(2020)Cornia, Stefanini, Baraldi, Corsini, and
  Cucchiara}]{cornia2020explaining}
Cornia M, Stefanini M, Baraldi L, Corsini M, Cucchiara R (2020) Explaining
  digital humanities by aligning images and textual descriptions. Pattern
  Recognition Letters 129:166--172

\bibitem[{Crowley and Zisserman(2014)}]{crowley2014search}
Crowley EJ, Zisserman A (2014) In search of art. In: European Conference on
  Computer Vision, Springer, pp 54--70

\bibitem[{Deng et~al(2009)Deng, Dong, Socher, Li, Li, and
  Fei-Fei}]{deng2009imagenet}
Deng J, Dong W, Socher R, Li LJ, Li K, Fei-Fei L (2009) Imagenet: A large-scale
  hierarchical image database. In: 2009 IEEE Conference on Computer Vision and
  Pattern Recognition, IEEE, pp 248--255

\bibitem[{Elgammal et~al(2017)Elgammal, Liu, Elhoseiny, and
  Mazzone}]{elgammal2017can}
Elgammal A, Liu B, Elhoseiny M, Mazzone M (2017) {CAN}: Creative adversarial
  networks, generating ``art'' by learning about styles and deviating from
  style norms. arXiv preprint arXiv:170607068

\bibitem[{Garcia and Vogiatzis(2018)}]{garcia2018read}
Garcia N, Vogiatzis G (2018) How to read paintings: Semantic art understanding
  with multi-modal retrieval. In: Proceedings of the European Conference on
  Computer Vision (ECCV)

\bibitem[{Garcia et~al(2020)Garcia, Renoust, and
  Nakashima}]{garcia2020contextnet}
Garcia N, Renoust B, Nakashima Y (2020) Context{N}et: Representation and
  exploration for painting classification and retrieval in context.
  International Journal of Multimedia Information Retrieval 9(1):17--30

\bibitem[{Gonthier et~al(2018)Gonthier, Gousseau, Ladjal, and
  Bonfait}]{gonthier2018weakly}
Gonthier N, Gousseau Y, Ladjal S, Bonfait O (2018) Weakly supervised object
  detection in artworks. In: Proceedings of the European Conference on Computer
  Vision (ECCV)

\bibitem[{Goodfellow et~al(2014)Goodfellow, Pouget-Abadie, Mirza, Xu,
  Warde-Farley, Ozair, Courville, and Bengio}]{goodfellow2014generative}
Goodfellow I, Pouget-Abadie J, Mirza M, Xu B, Warde-Farley D, Ozair S,
  Courville A, Bengio Y (2014) Generative adversarial nets. In: Advances in
  Neural Information Processing Systems, pp 2672--2680

\bibitem[{Gultepe et~al(2018)Gultepe, Conturo, and
  Makrehchi}]{gultepe2018predicting}
Gultepe E, Conturo TE, Makrehchi M (2018) Predicting and grouping digitized
  paintings by style using unsupervised feature learning. Journal of Cultural
  Heritage 31:13--23

\bibitem[{Guo et~al(2017)Guo, Liu, Zhu, and Yin}]{guo2017deep}
Guo X, Liu X, Zhu E, Yin J (2017) Deep clustering with convolutional
  autoencoders. In: International Conference on Neural Information Processing,
  Springer, pp 373--382

\bibitem[{He et~al(2016)He, Zhang, Ren, and Sun}]{he2016deep}
He K, Zhang X, Ren S, Sun J (2016) Deep residual learning for image
  recognition. In: Proceedings of the IEEE Conference on Computer Vision and
  Pattern Recognition, pp 770--778

\bibitem[{Hogan et~al(2021)Hogan, Blomqvist, Cochez, d’Amato, Melo,
  Gutierrez, Kirrane, Gayo, Navigli, Neumaier et~al}]{hogan2021knowledge}
Hogan A, Blomqvist E, Cochez M, d’Amato C, Melo GD, Gutierrez C, Kirrane S,
  Gayo JEL, Navigli R, Neumaier S, et~al (2021) Knowledge graphs. ACM Computing
  Surveys (CSUR) 54(4):1--37

\bibitem[{Huang et~al(2017)Huang, Liu, Van Der~Maaten, and
  Weinberger}]{huang2017densely}
Huang G, Liu Z, Van Der~Maaten L, Weinberger KQ (2017) Densely connected
  convolutional networks. In: Proceedings of the IEEE Conference on Computer
  Vision and Pattern Recognition, pp 4700--4708

\bibitem[{Karayev et~al(2013)Karayev, Trentacoste, Han, Agarwala, Darrell,
  Hertzmann, and Winnemoeller}]{karayev2013recognizing}
Karayev S, Trentacoste M, Han H, Agarwala A, Darrell T, Hertzmann A,
  Winnemoeller H (2013) Recognizing image style. arXiv preprint arXiv:13113715

\bibitem[{Khan et~al(2014)Khan, Beigpour, Van~de Weijer, and
  Felsberg}]{khan2014painting}
Khan FS, Beigpour S, Van~de Weijer J, Felsberg M (2014) Painting-91: A large
  scale database for computational painting categorization. Machine Vision and
  Applications 25(6):1385--1397

\bibitem[{Krizhevsky et~al(2012)Krizhevsky, Sutskever, and
  Hinton}]{krizhevsky2012imagenet}
Krizhevsky A, Sutskever I, Hinton GE (2012) Imagenet classification with deep
  convolutional neural networks. In: Advances in Neural Information Processing
  Systems, pp 1097--1105

\bibitem[{LeCun et~al(1989)LeCun, Boser, Denker, Henderson, Howard, Hubbard,
  and Jackel}]{lecun1989backpropagation}
LeCun Y, Boser B, Denker JS, Henderson D, Howard RE, Hubbard W, Jackel LD
  (1989) Backpropagation applied to handwritten zip code recognition. Neural
  computation 1(4):541--551

\bibitem[{LeCun et~al(2015)LeCun, Bengio, and Hinton}]{lecun2015deep}
LeCun Y, Bengio Y, Hinton G (2015) Deep learning. Nature 521(7553):436--444

\bibitem[{Leder et~al(2004)Leder, Belke, Oeberst, and
  Augustin}]{leder2004model}
Leder H, Belke B, Oeberst A, Augustin D (2004) A model of aesthetic
  appreciation and aesthetic judgments. British Journal of Psychology
  95(4):489--508

\bibitem[{Lu et~al(2019)Lu, Duan, and Zhang}]{lu2019audio}
Lu R, Duan Z, Zhang C (2019) Audio--visual deep clustering for speech
  separation. IEEE/ACM Transactions on Audio, Speech, and Language Processing
  27(11):1697--1712

\bibitem[{Van~der Maaten and Hinton(2008)}]{van2008visualizing}
Van~der Maaten L, Hinton G (2008) Visualizing data using t-{SNE}. Journal of
  Machine Learning Research 9(11)

\bibitem[{McInnes et~al(2018)McInnes, Healy, and Melville}]{mcinnes2018umap}
McInnes L, Healy J, Melville J (2018) {UMAP}: Uniform manifold approximation
  and projection for dimension reduction. arXiv preprint arXiv:180203426

\bibitem[{Ren et~al(2019)Ren, Hu, Dai, Pan, Hoi, and Xu}]{ren2019semi}
Ren Y, Hu K, Dai X, Pan L, Hoi SC, Xu Z (2019) Semi-supervised deep embedded
  clustering. Neurocomputing 325:121--130

\bibitem[{Rousseeuw(1987)}]{rousseeuw1987silhouettes}
Rousseeuw PJ (1987) Silhouettes: A graphical aid to the interpretation and
  validation of cluster analysis. Journal of Computational and Applied
  Mathematics 20:53--65

\bibitem[{Saleh et~al(2016)Saleh, Abe, Arora, and Elgammal}]{saleh2016toward}
Saleh B, Abe K, Arora RS, Elgammal A (2016) Toward automated discovery of
  artistic influence. Multimedia Tools and Applications 75(7):3565--3591

\bibitem[{Shamir et~al(2010)Shamir, Macura, Orlov, Eckley, and
  Goldberg}]{shamir2010impressionism}
Shamir L, Macura T, Orlov N, Eckley DM, Goldberg IG (2010) Impressionism,
  expressionism, surrealism: Automated recognition of painters and schools of
  art. ACM Transactions on Applied Perception (TAP) 7(2):8

\bibitem[{Shen et~al(2019)Shen, Efros, and Mathieu}]{shen2019discovering}
Shen X, Efros AA, Mathieu A (2019) Discovering visual patterns in art
  collections with spatially-consistent feature learning. arXiv preprint
  arXiv:190302678

\bibitem[{Simonyan and Zisserman(2014)}]{simonyan2014very}
Simonyan K, Zisserman A (2014) Very deep convolutional networks for large-scale
  image recognition. arXiv preprint arXiv:14091556

\bibitem[{Spehr et~al(2009)Spehr, Wallraven, and Fleming}]{spehr2009image}
Spehr M, Wallraven C, Fleming RW (2009) Image statistics for clustering
  paintings according to their visual appearance. In: Computational Aesthetics
  2009: Eurographics Workshop on Computational Aesthetics in Graphics,
  Visualization and Imaging, Eurographics, pp 57--64

\bibitem[{Strezoski and Worring(2017)}]{strezoski2017omniart}
Strezoski G, Worring M (2017) Omni{A}rt: Multi-task deep learning for artistic
  data analysis. arXiv preprint arXiv:170800684

\bibitem[{Tan et~al(2016)Tan, Chan, Aguirre, and Tanaka}]{tan2016ceci}
Tan WR, Chan CS, Aguirre HE, Tanaka K (2016) Ceci n'est pas une pipe: A deep
  convolutional network for fine-art paintings classification. In: 2016 IEEE
  International Conference on Image Processing (ICIP), IEEE, pp 3703--3707

\bibitem[{Tan et~al(2018)Tan, Chan, Aguirre, and Tanaka}]{tan2018improved}
Tan WR, Chan CS, Aguirre HE, Tanaka K (2018) Improved {A}rt{GAN} for
  conditional synthesis of natural image and artwork. IEEE Transactions on
  Image Processing 28(1):394--409

\bibitem[{Tomei et~al(2019)Tomei, Cornia, Baraldi, and
  Cucchiara}]{tomei2019art2real}
Tomei M, Cornia M, Baraldi L, Cucchiara R (2019) Art2{R}eal: Unfolding the
  reality of artworks via semantically-aware image-to-image translation. In:
  Proceedings of the IEEE Conference on Computer Vision and Pattern
  Recognition, pp 5849--5859

\bibitem[{Vaigh et~al(2021)Vaigh, Garcia, Renoust, Chu, Nakashima, and
  Nagahara}]{vaigh2021gcnboost}
Vaigh CBE, Garcia N, Renoust B, Chu C, Nakashima Y, Nagahara H (2021)
  {GCNB}oost: Artwork classification by label propagation through a knowledge
  graph. arXiv preprint arXiv:210511852

\bibitem[{Van~Noord et~al(2015)Van~Noord, Hendriks, and Postma}]{van2015toward}
Van~Noord N, Hendriks E, Postma E (2015) Toward discovery of the artist's
  style: Learning to recognize artists by their artworks. IEEE Signal
  Processing Magazine 32(4):46--54

\bibitem[{Vinh et~al(2010)Vinh, Epps, and Bailey}]{vinh2010information}
Vinh NX, Epps J, Bailey J (2010) Information theoretic measures for clusterings
  comparison: Variants, properties, normalization and correction for chance.
  The Journal of Machine Learning Research 11:2837--2854

\bibitem[{Westlake et~al(2016)Westlake, Cai, and Hall}]{westlake2016detecting}
Westlake N, Cai H, Hall P (2016) Detecting people in artwork with {CNN}s. In:
  European Conference on Computer Vision, Springer, pp 825--841

\bibitem[{Xie et~al(2016)Xie, Girshick, and Farhadi}]{xie2016unsupervised}
Xie J, Girshick R, Farhadi A (2016) Unsupervised deep embedding for clustering
  analysis. In: International Conference on Machine Learning, PMLR, pp 478--487

\bibitem[{Yang et~al(2017)Yang, Fu, Sidiropoulos, and Hong}]{yang2017towards}
Yang B, Fu X, Sidiropoulos ND, Hong M (2017) Towards k-means-friendly spaces:
  Simultaneous deep learning and clustering. In: Proceedings of the 34th
  International Conference on Machine Learning-Volume 70, JMLR. org, pp
  3861--3870

\end{thebibliography}

%
%

\end{document}